\documentclass{ecai}
\usepackage{times}
\usepackage{graphicx}
\usepackage{latexsym}
\usepackage{footnote}

\usepackage[utf8]{inputenc}
\DeclareUnicodeCharacter{00B0}{\ensuremath{^\circ}}

\usepackage{amsmath}
\usepackage{booktabs}
\usepackage{xcolor}

\newcommand{\fausto}[1]{\textcolor{red}{\textbf{[Fausto: #1]}}}

\title{Multi-Modal Subjective Context Modelling and Recognition}
\author{
Qiang Shen $^{1,2}$

\and Stefano Teso$^{2}$ 
\and Wanyi Zhang$^{2}$ 

\and Hao Xu $^{1}$

\and Fausto Giunchiglia $^{1,2}$ 

}

\begin{document}
\maketitle

\footnotetext[1]{College of Computer Science and Technology, Jilin University, Changchun, China, email: shenqiang19@mails.jlu.edu.cn, xuhao@jlu.edu.cn}
\footnotetext[2]{University of Trento, Italy, email: \{stefano.teso, wanyi.zhang, \\fausto.giunchiglia \}@unitn.it}

\begin{abstract}
    Applications like personal assistants need to be aware of the user's context, e.g., where they are, what they are doing, and with whom.
    Context information is usually inferred from sensor data, like GPS sensors and accelerometers on the user's smartphone.  This prediction task is known as context recognition.  A well-defined context model is fundamental for successful recognition.
    Existing models, however, have two major limitations.  First, they focus on few aspects, like location or activity, meaning that recognition methods based on them can only compute and leverage few inter-aspect correlations.  Second, existing models typically assume that context is objective, whereas in most applications context is best viewed from the user's perspective.  Neglecting these factors limits the usefulness of the context model and hinders recognition.
    We present a novel ontological context model that captures \emph{five dimensions}, namely time, location, activity, social relation and object.  Moreover, our model defines \emph{three levels of description} (objective context, machine context, and subjective context) that naturally support subjective annotations and reasoning.
    An initial context recognition experiment on real-world data hints at the promise of our model.
\end{abstract}

\section{INTRODUCTION}

The term ``context'' refers to any kind of information necessary to describe the situation that an individual is in~\cite{dey2001understanding}.
Automatic recognition of personal context is the key in applications like personal assistants, smart environments, and health monitoring apps, because it enables intelligent agents to respond proactively and appropriately based on (an estimate of) their user's context.
For instance, a personal assistant aware that its user is at home, alone, doing housework, could suggest him or her to order a take-away lunch.
Since context information is usually not available, the machine has to infer it from sensor data, like GPS coordinates, acceleration, and nearby Bluetooth devices measured by the user's smartphone.  The standard approach to \emph{context recognition} is to train a machine learning model on a large set of sensor readings and corresponding context annotations to predict the latter from the former.  Existing implementations are quite diverse, and range from shallow models like logistic regression~\cite{vaizman2017recognizing} to deep neural networks like feed-forward networks~\cite{vaizman2018context}, LSTMs~\cite{hammerla2016deep}, and CNNs~\cite{saeed2018learning}. 

A context model defines how context data are structured.  A good context model should capture all kinds of situational information relevant to the application at hand~\cite{dey2001understanding} and use the right level of abstraction~\cite{bettini2010survey}.  Ontology is a widely accepted tool for formalizing context information~\cite{krummenacher2007ontology}, and several context ontologies have been proposed.  Typical examples include CONON~\cite{wang2004ontology} and CaCONT~\cite{xu2013cacont}.  CONON focuses on modeling locations by providing an upper ontology and lower domain-specific ontologies organized into a hierarchy.  CaCONT defines several types of entities, and provides different levels of abstraction for specifying location of entities, e.g., GPS and location hierarchies.  Focusing on semantic information of place, the work in~\cite{zavala2015platys} proposed a place-oriented ontology model representing different levels of place and related activities and improve the performance of place recognition.  In~\cite{kola2019s}, they proposed an ontology model involving social situation and the interaction between people.

These models, however, suffer from two main limitations.
First, in order to support context recognition, the model should account for subjectivity of context descriptions.  For instance, the \emph{objective} location ``hospital'' plays different roles for different people:  for patients it is a ``place for recovering'', while for nurses it is a ``work place''.  This makes all the difference for personal assistants because the services that a user needs strongly depend on his or her subjective viewpoint.  Most context models ignore this fact, with few exceptions, cf.~\cite{kokar2009ontology}.
Second, arguably answers to five basic questions -- ``what time is it?'', ``where are you?'', ``what are you doing?'',  ``who are you with?'', and  ``what are you with?''-- are necessary to define human contexts.  Correlations between these aspects are also fundamental in recognition and reasoning:  if the user is in her room, a personal assistant should be more likely to guess that she is ``studying'' or ``resting'', rather than ``swimming''.  In stark contrast, most models
are restricted to one or few of the above five aspects and therefore fail to capture important correlations, like those between activity and location or between time and social context. 


As a remedy, we introduce a novel ontological context model that
supports both reasoning and recognition from a subjective perspective,
that captures time, location, activity, social relations and object,
and and that enables downstream context recognition tools to leverage correlations between these five fundamental dimensions.  Our model also incorporates three levels of description for each aspect, namely objective, machine-level, and subjective, which naturally support different kinds of annotations.  We apply and test our approach by collaborating with sociology experts within the SmartUnitn-One project~\cite{giunchiglia2018mobile}.  We validate empirically our model by evaluating context recognition performance on the SmartUnitn-One context and sensor annotation data set~\cite{giunchiglia2018mobile}, which was annotated consistently with our context model.  Our initial results shows that handling correlations across aspects substantially improves recognition performance and makes it possible to predict activities that are otherwise very hard to recognize.

\section{CONTEXT MODELLING}

Context is a theory of the world that encodes an individual' subjective perspective about it~\cite{giunchiglia1993contextual}.  Individuals have a limited and partial view of the world at all times in their everyday life. For instance, consider a classroom with a teacher and a few students.  Despite all the commonalities, each person in the room has a different context because they focus on different elements of their personal experience (the students focus on the teacher while the teacher focuses on the students) and ignore others (like the sound of the projector, the weather outside, and so on.)  Given the diversity and complexity of individual experiences, formalizing the notion of context in its entirety is essentially impossible.  For this reason, simpler but useful application-specific solutions are necessary.

\begin{figure}[tb]
    \centering
    \includegraphics[width=0.98\linewidth]{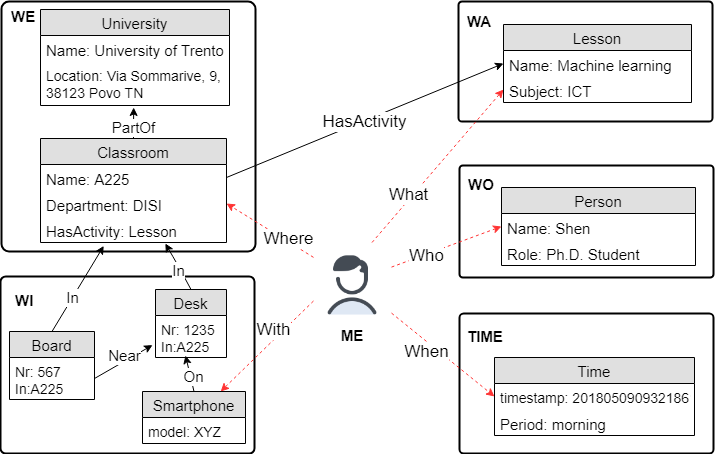}
    \caption{Illustration of our context model.}
    \label{fig:context_model}
\end{figure}

Previous work has observed that reasoning in terms of questions like ``what time is it?'', ``where are you?'', ``what are you doing?'', ``who are you with?'', ``what are you with?'' is fundamental for describing and collecting the behavior of individuals~\cite{giunchiglia1993contextual}.  Motivated by this observation and our previous work~\cite{2017-Giunchiglia,giunchiglia2017personal,2020-Osman} , we designed an ontology-based context model organized according to the aforementioned dimensions of the world:  time, location, activity, social relations and object.  Formally, context is defined as a tuple:
$$
    \text{Context} = \langle \text{TIME}, \text{WE}, \text{WA}, \text{WO}, \text{WI} \rangle
$$
where:
\begin{description}

    \item[TIME] captures the exact time of context, e.g., ``morning''.  We refer to it as the \emph{temporal context}.  Informally, it answers the question ``When did this context occur?''.

    \item[WE] captures the exact location of context, e.g., ``classroom''.  We refer to it as the \emph{endurant context}.  Informally, it answers the question ``Where are you?''.

    \item[WA] captures the activity of context, e.g., ``studying''.  We refer to it as the \emph{perdurant context}.  Informally, it answers the question ``What are you doing?''.

    \item[WO] captures the social relations of context, e.g., ``friend''.  We refer to it as the \emph{social context}.  Informally, it answers the question ``Who are you with?''.

    \item[WI] captures the materiality of context, e.g., ``smartphone''. We refer to it as the \emph{object context}. Informally, it answers the question ``What are you with?''.

\end{description}
Figure~\ref{fig:context_model} shows a scenario as a knowledge graph representing the personal context of an individual in the class. For instance, attributes of WO are “Class”, “Name”, and “Role”, and their values are “Person”, “Shen”, and “PhD student”, respectively. Edges represent relations between entities, e.g., “Shen” is in relation “Attend” with “Lesson”.  

The example in Figure~\ref{fig:context_model} is presented in objective terms, that is, facts are stated as if they were independent of personal conscious experiences.  However, each person interprets the world and her surroundings from her personal privileged point of view, which accounts for her personal knowledge, mental characteristics, states, etc.  For instance, while in Figure~\ref{fig:context_model} ``Shen'' has an objective role of Ph.D student, for other people ``Shen'' plays the roles of a ``friend'' or a “classmate” subjectively.  The subjective context which is related to personal consciousness, knowledge, etc. can provide more information for applications such as personal assistant in order to give more intelligent services. 

Notice that a person's view of her context is radically different from what her handheld personal assistant observes.  In fact, machines interpret the world via sensors, while humans do not only interpret the world via their perceptions but with their knowledge as well.  For instance, while a machine views location (e.g., a building) as a set of coordinates, humans interpret it based on its \emph{function} (e.g., whether the building is their home or office).

\begin{table*}[tb]
    \centering

    \begin{small}
    \begin{tabular}{lllll}
        \toprule
        Level  & TIME & WE & WA & WO  \\
        \midrule
        Objective Context  &  2020-02-17
        11am & Via Sommarive, 9,
        38123 Povo TN  & Lesson & Shen \\
        Machine Context & 1581938718026 & 46°04'01.9"N 11°09'02.4"E & Accelerometer: 0g,0g,0g & “Shen” is in contact list \\
        Subjective Context & Morning & Classroom & Studying & Friend \\
        \bottomrule
    \end{tabular}
    \end{small}

    \caption{An example of our three-partitioned context model.  Each row gives a different description of the same underlying situation from the perspective of the world (top), the machine (middle), and the user (bottom).}
    \label{tab:context_example}
\end{table*}

\begin{figure}[tb]
    \centering
    \includegraphics[width=0.96\linewidth]{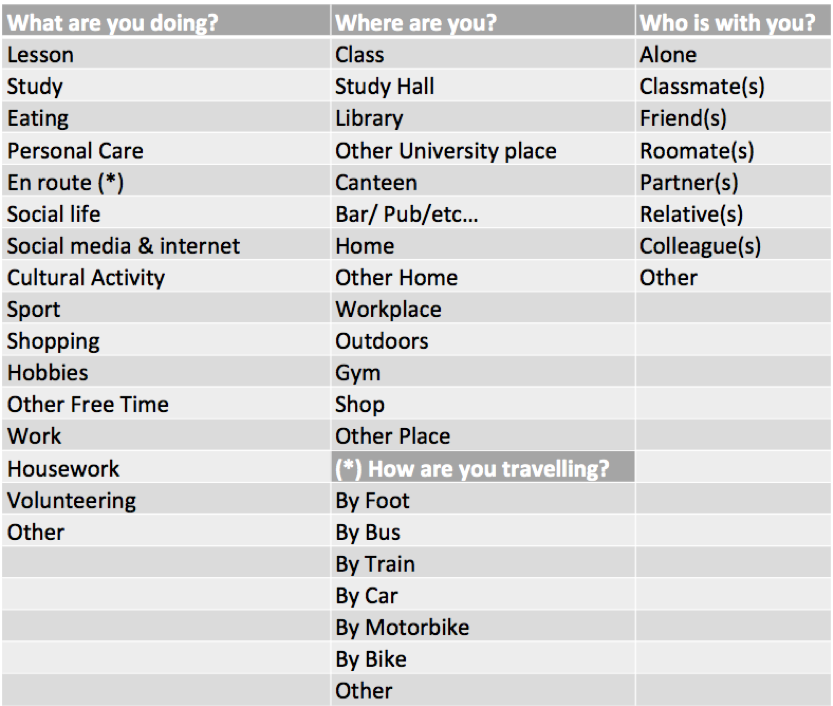}
    \caption{Questions and answers in the SmartUnitn-One questionnaire.}
    \label{fig:time-diary}
\end{figure}

\begin{table}
    \centering
    \begin{footnotesize}
    \begin{tabular}{p{12em}ll}
        \toprule
        Sensor              & Frequency         & Unit  \\
        \midrule
        Acceleration        & 20 Hz               & $m/s^2$              \\
        Linear Acceleration & 20 Hz               & $m/s^2$              \\
        Gyroscope           & 20 Hz               & rad$/s$             \\
        Gravity             & 20 Hz               & $m/s^2$              \\
        Rotation Vector     & 20 Hz               & Unitless          \\
        Magnetic Field      & 20 Hz               & $\mu T$             \\
        Orientation         & 20 Hz               & Degrees           \\
        Temperature         & 20 Hz               & °C                \\
        Atmospheric Pressure& 20 Hz               & hPa               \\
        Humidity            & 20 Hz               & \%                \\
        Proximity           & On change           & 0/1               \\
        Position            & Every minute        & Lat.$/$Lon. \\
        WIFI Network Connected & On change        & Unitless          \\
        WIFI Networks Available & Every minute    & Unitless       \\
        Running Application    & Every 5 seconds  & Unitless          \\
        Battery Level       & On change           & \%        \\
        Audio from the internal mic & 10 seconds per minute & Unitless \\
        Notifications received & On change        & Unitless          \\
        Touch event         & On change           & 0/1               \\
        Cellular network info & Once per minute & Unitless\\
        Screen Status, Flight Mode, Battery Charge, Doze Mode, Headset Plugged in, Audio Mode, Music Playback       & On change           & 0/1               \\
        \bottomrule
    \end{tabular}
    \end{footnotesize}
    \caption{List of sensors.  Proximity triggers when the phone detects very
    close objects, e.g., the user's ear during a phone call.}
    \label{tab:sensors}
\end{table}

To model context precisely and completely, in addition to considering five dimensions, as discussed above, we also model three perspectives:  objective context, subjective context and machine context.  Table~\ref{tab:context_example} shows the above example viewed through three types of perspective.  The objective context captures the fact that at the University of Trento, Italy, at 11:00 AM, a person is attending a class together with Shen.  When moving from objective to subjective, things change dramatically.  From the perspective of the machine, the temporal context ``11:00 AM'' is viewed as a timestamp timestamp ``1581938718026'', and in subjective terms it becomes ``morning''; similarly, ``University of Trento” becomes coordinates “46°04'N,11°09'E” for the machine and “classroom” from a subjective perspective.  For the perdurant context, the activity of taking lesson can be subjectively annotated as “study” by user, but it can be described as “connecting WIFI of classroom, sensors such as gyroscope, accelerometer are sensed as static”. For the social context, “Shen” is described as friend subjectively by the user and the machine senses “Shen” is in the contact list of the user.

\section{EMPIRICAL EVALUATION}

In order to evaluate the proposed context model, we carried out a context recognition experiment using the SmartUnitn-One data set~\cite{giunchiglia2018mobile}, and studied whether recognition of subjective context is feasible and whether taking inter-aspect correlations into account helps recognition performance. \\

\noindent \textbf{Data Collection.}  The SmartUnitn-One data set consists of sensor readings and context annotations obtained from 72 volunteers (university students) for a period of two weeks. All participants were required to install the i-Log app~\cite{zeni2014multi}, which simultaneously records sensor data from several sensors (cf. Table~\ref{tab:sensors}) and context annotations.   During the first week, students were asked to report their own context every 30 minutes by administering them questionnaires comprising three questions about location, activity, and social relations.  The i-Log app collected sensor data at the same time.  During the second week, the participants were only required to have the application running for the sensor data collection.  All records were timestamped automatically.  The questions were designed according to our context model and possible answers were modelled following the America Time Use Survey (ATUS)~\cite{shelley2005developing}, leading to an ontology with over 80 candidate labels, see Figure~\ref{fig:time-diary} for the full list.  Object context (WI) information was not collected as it is too hard to track without disrupting the volunteer's routines.  All records were processed as in~\cite{zeni2019fixing}.  This resulted in 23309 records, each comprising 122 sensor readings (henceforth, features) and self-reported annotations about location, activity, and social context. \\

\noindent \textbf{Experimental Setup.}  For every aspect in $\{\text{WA}, \text{WE}, \text{WO}\}$, we trained a random forest to predict that aspect from sensor measurements.  We randomly split the dataset into training ($75\%$ of the records) and validation ($25\%$ of the records) subsets and then selected the maximum depth of the forest using the validation set only.  The classifier performance was evaluated 
using a rigorous $5$-fold cross validation procedure.  The data set was randomly partitioned into 5 folds.  We hold out the selected fold as the test set to train a classifier on the remaining folds and compute the performance on the held out (test) fold.  Then, we compared this model to another random forests (with the same maximum depth) that was supplied both sensor data and annotations for (a subset of) the other aspects as inputs.  In order to account for label skew (e.g., some locations and activities are much more frequent than others), performance was measured using the \emph{micro-average} $F_1$ score to account for class imbalance.\\

\begin{figure*}[tb]
    \centering
    \begin{tabular}{ccc}
        \includegraphics[width=0.3\linewidth]{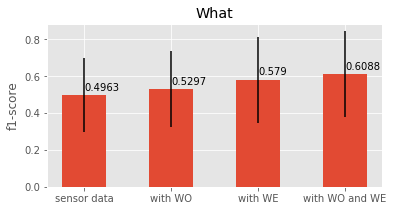} &
        \includegraphics[width=0.3\linewidth]{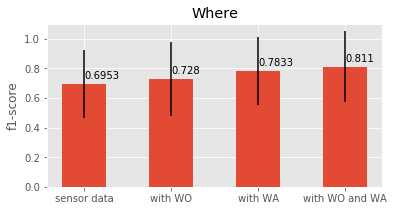} &
        \includegraphics[width=0.3\linewidth]{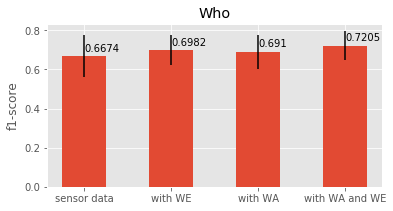}
    \end{tabular}
    \caption{$F_1$ of our context recognition model.  From left to right: perdurant (WA), endurant (WE), and social context (WO), respectively.  The leftmost column refers to a predictor that uses sensor data only, while the other columns to predictors that in addition have access to context annotations.}
    \label{fig:results}
\end{figure*}

\noindent \textbf{Results and Discussion.}  The average $F_1$ score across users are reported in Figure~\ref{fig:results}.  The plots show very clearly that knowledge of other aspects substantially improves  recognition performance regardless of the aspect being predicted:  supplying the other aspects as inputs increases the $F_1$ score of predicting WA and WE by more than $10\%$ and for WO by more than $5\%$.  A breakdown of performance increase can be viewed in Table~\ref{tab:improvemetn}.
The table shows that all aspects are correlated, as expected, especially activity and location, and that providing more aspects as inputs increases $F_1$ almost additively.

\begin{table}
    \centering
    \begin{footnotesize}
    \begin{tabular}{p{12em}lll}
        \toprule
        Inputs                  & WA        & WE        & WO        \\
        \midrule
        Sensors + WA            & --        & $+8.80\%$    & $+2.36\%$    \\
        Sensors + WE            & $+8.27\%$    & --        & $+3.09\%$    \\
        Sensors + WO            & $+3.34\%$    & $+3.27\%$    & --        \\
        Sensors + Other Aspects & $+11.25\%$   & $+11.57\%$   & $+5.31\%$    \\
        \bottomrule
    \end{tabular}
    \end{footnotesize}
    \caption{Improvement in $F_1$ score when using other aspects as inputs to the recognition model.  Columns indicate the aspect being predicted.}
    \label{tab:improvemetn}
\end{table}

\begin{figure*}[tb]
    \centering
    \begin{tabular}{ccc}
        \includegraphics[width=0.98\linewidth]{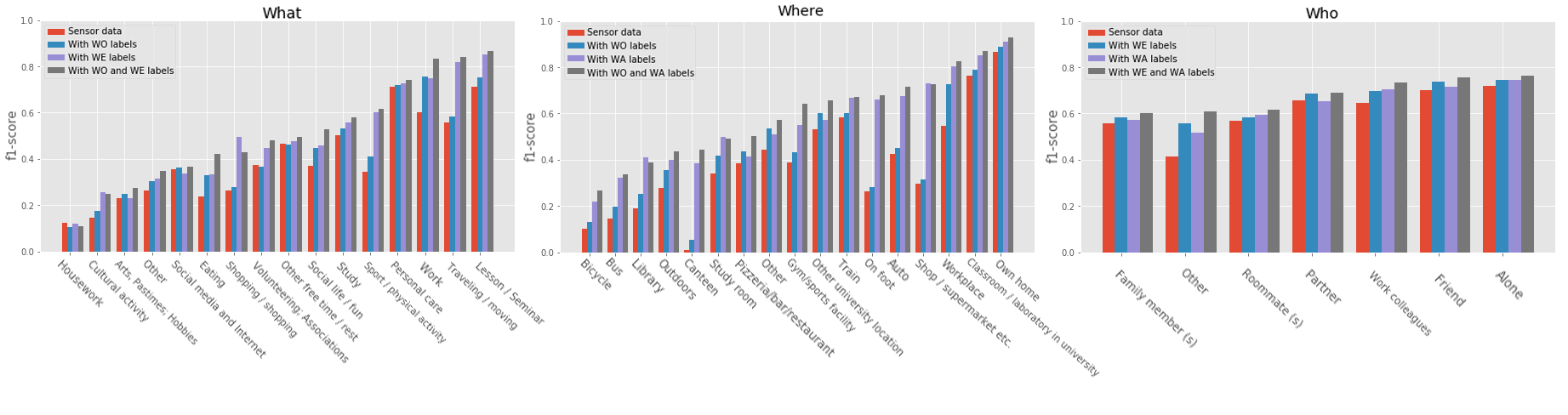}
    \end{tabular}
    \caption{$F_1$ of individual labels (averaged over users).  From left to right: perdurant, endurant, and social context, respectively.}
    \label{fig:recognition}
\end{figure*}

\noindent Figure~\ref{fig:recognition} shows $F_1$ scores (again, averaged across users) for each label.
For \textbf{WO}, some labels are clearly easier to predict than others. The performance improvement is usually in the $5$--$10\%$ range, with the notable exception of ``other'', which improves by about $20\%$.  It seems that location information always facilitates recognition of WO, while activity does not.  Their combination, however, is always beneficial.
For \textbf{WE}, looking at either WO and WA helps recognition performance in all cases, and providing both WO and WA gives a larger improvement than than providing them separately.  The exceptions are ``library'', ``study room'', and ``shop'', for which knowing WA improves more than knowing both WO and WA.  This is somewhat surprising, as we expect social context to be moderately indicative of location, and deserves further investigation.  Some locations (``canteen'', ``on foot'', ``auto'', ``shop'', and ``workplace'') receive a major increase in recognition performance, from $25\%$ to $40\%$ approximately.  This is partly due to the rarity of these classes in the data set, which shows that inter-aspect correlations supply to the lack of supervision.
Finally for \textbf{WA}, some activities (like ``housework'', ``cultural activities'', and ``hobbies'') are very hard to predict, as their $F_1$ score is below $30\%$, while others (``work'', ``moving'', and ``lesson'') are much easier to predict, with more than $80\%$ $F_1$ score.  This mostly shows that rare activities are harder to predict, understandably, although other factors might play a role.  Using the full context (with WE and WO) always improves performance, except for ``housework''. For all the other activities, the improvement is from $5\%$ to $20\%$, and even larger for ``Shopping'', ``Sport'' and ``Traveling'', for which the improvement is up to $30\%$.

This analysis provides ample support for our context model: correlations between different aspects improve context recognition performance for most users and, even more importantly, some values (like ``Canteen'') that are essentially impossible to recognize suddenly become much easier when full context information is provided.

\section{CONCLUSION}

We designed a novel context model that captures situational information about time, location, activity, and social relations of individuals using subjective---rather than objective---terms.  An initial context recognition experiments on real-world data showed that machine learning models built using our context model produce higher quality predictions than models based on less complete context models.  As for future work, we plan to study the effects of subjectivity more in detail, to migrate our architecture to more refined learning approaches (e.g., deep neural nets), and to carry out an extensive comparison against the state-of-the-art in context recognition.

\section{ACKNOWLEDGEMENT}

The research of FG has received funding from the European Union's Horizon 2020 FET Proactive project ``WeNet -- The Internet of us'', grant agreement No 823783.  The research of ST and WZ have received funding from the ``DELPhi - DiscovEring Life Patterns'' project funded by the MIUR Progetti di Ricerca di Rilevante Interesse Nazionale (PRIN) 2017 -- DD n. 1062 del 31.05.2019.

\bibliographystyle{ecai}
\end{document}